\documentclass[letterpaper]{article} 
\usepackage{aaai2027}  
\usepackage[hyphens]{url}  
\usepackage{graphicx} 
\usepackage{natbib}  
\usepackage{caption} 
\usepackage{amsmath}
\usepackage{amssymb}
\usepackage{algorithm}
\usepackage{algorithmic}
\usepackage{amsmath}
\usepackage[table]{xcolor} 
\usepackage{newfloat}
\usepackage{listings}
\DeclareCaptionStyle{ruled}{labelfont=normalfont,labelsep=colon,strut=off} 
\floatstyle{ruled}
\newfloat{listing}{tb}{lst}{}
\floatname{listing}{Listing}

\usepackage{booktabs}

\title{ECG-LENS: Lead-Aware Clinical Context Enriched ECG Report Generation and Evaluation}
\author{
    Akanta Das\textsuperscript{\rm 1},
    Tasinul Islam Ahon\textsuperscript{\rm 1},
    Ahmed Mahir Sultan Rumi\textsuperscript{\rm 1},\\
    Md Mahbubur Rahman\textsuperscript{\rm 2},
    Tausif Amim Shadly\textsuperscript{\rm 3},
    Tanzima Hashem\textsuperscript{\rm 1}
}
\affiliations{
    \textsuperscript{\rm 1}Bangladesh University of Engineering and Technology, Dhaka, Bangladesh\\
    \textsuperscript{\rm 2}Samsung Research America, USA\\
    \textsuperscript{\rm 3}National Health Service, UK
}

\begin{document}

\maketitle

\begin{abstract}

Electrocardiography (ECG) is one of the most widely used non-invasive tools for diagnosing cardiovascular disease, but transforming multi-lead ECG recordings into reliable clinical reports remains challenging. Automating ECG report generation could reduce clinicians’ interpretive workload, improve diagnostic efficiency, and expand access to cardiac assessment in rural and underserved communities. Unlike image-based report generation tasks such as chest X-ray reporting, ECG interpretation requires the analysis of subtle temporal morphologies, lead-specific abnormalities, and inter-lead relationships, followed by coherent diagnostic reasoning expressed in dense clinical terminology. Existing systems predominantly focus on classification, while current report-generation methods often produce outputs that remain inadequate for practical clinical use. To address these challenges, we propose ECG-LENS, an end-to-end ECG report generation framework that jointly models multi-lead signals, diagnosis-aware representations, and clinically grounded report generation. ECG-LENS combines lead-wise encoders that preserve localized waveform morphology with a global encoder that captures inter-lead dependencies. To guide report generation, we fuse signal representations with clinically enriched textual prompts that condition a GPT-2 decoder. We further introduce an ECG-specific report preprocessing strategy that removes repetitive and non-informative text, encouraging the model to focus on clinically meaningful findings. Because clinically correct ECG reports can differ in wording, lexical metrics alone may under- or overestimate report quality. We therefore propose F1-ECGBERT, a BERT-based ECG-specific metric that measures agreement between diagnostic labels extracted from generated and reference reports. Experiments on PTB-XL and cross-domain evaluation on MIMIC-IV-ECG show that ECG-LENS consistently outperforms state-of-the-art methods, with absolute gains of 4.0\%, 6.3\%, and 11.5\% in METEOR, ROUGE-L, and F1-ECGBERT, respectively, over the strongest baselines.


\end{abstract}


\section{Introduction}
\label{sec:introduction}

Cardiovascular diseases (CVDs) remain among the most serious public health burdens worldwide, making timely and accessible cardiac assessment essential for preventing avoidable complications. Electrocardiography (ECG) is central to this effort because it is non-invasive, inexpensive, rapid, and routinely available. Although ECG acquisition is straightforward, its interpretation requires
substantial expertise. A standard 12-lead ECG is a multivariate time series in which the limb and precordial leads observe cardiac electrical activity from complementary perspectives. Diagnostic evidence is therefore encoded not only in temporal morphology, such as rhythm, intervals, and ST/T-wave patterns, but also in the spatial distribution of abnormalities across specific leads. A cardiologist or trained physician must inspect waveform morphology, integrate evidence across leads, and translate the findings into a concise diagnostic report. This process is time-consuming and access to specialists may be limited in rural and underprivileged communities. Automated ECG report generation could consequently reduce reporting workload and broaden access to cardiac assessment. 


Deep learning has achieved strong performance in automated ECG analysis,
particularly in disease classification ~\citep{hannun2019,ribeiro2020,strodthoff2021}. More recent research has progressed from self-supervised ECG representation learning and ECG--text alignment~\citep{na2024,jin2025,liu2024} to
multimodal and instruction-tuned language models for report
generation~\citep{wan2025,qiang2025,xia2025,yang2026}. However, classification produces isolated labels rather than an integrated account of rhythm, axis, conduction, ST/T
changes, affected leads, and overall interpretation required in clinical
reporting. Moreover, although automated report generation has achieved substantial progress in
image-based domains such as radiology, comparable advances in ECG report
generation remain limited. Unlike medical images, ECG recordings are
multivariate time-series signals whose interpretation requires jointly
modeling subtle temporal morphology and lead-specific relationships. This
inherent complexity contributes to the comparatively limited clinical fidelity of
existing ECG report-generation systems and also limits the direct transfer of radiology-report generation methods.

Despite recent progress, major limitations remain.
First, most ECG--language systems compress the complete recording into a single representation before decoding, suffering \emph{loss of lead-specific evidence}. Because different leads provide distinct electrical views, early aggregation can dilute localized evidence; for example, ST elevation in leads II, III, and aVF conveys a specific diagnostic pattern indicative of inferior myocardial infarction rather than a generic global abnormality~\citep{thygesen2018,wagner2009}. Second, directly conditioning a language model on a global signal embedding provides no explicit diagnostic plan, while abbreviated, multilingual, and inconsistently phrased reference reports introduce ambiguity into the learning targets. It suffers from 
\emph{weak clinical grounding and supervision}. Finally, traditional metric like BLEU, ROUGE, and METEOR quantify lexical overlap but cannot reliably distinguish valid paraphrases from fluent reports that omit, alter, or hallucinate clinically important findings. It deems \emph{clinically incomplete evaluation.} Evaluation should therefore measure preservation of diagnostic content in addition to surface similarity.

To address these limitations, we propose ECG-LENS (Lead-aware Enrichment with Context Narrative Synthesis), a lightweight lead-aware ECG report generation framework with clinical context enrichment. ECG-LENS preserves individual-lead morphology through lead-specific encoders, models global inter-lead relationships using a global encoder, constructs a clinical context-enriched text prompt as an intermediate clinical representation to guide a GPT-2 text decoder. Our main contributions are:

\begin{itemize}

    \item \textbf{Lead-aware local and global representation learning.}
    We combine lead-wise ResNet-18 encoders that retain localized waveform morphology with a global encoder that captures inter-lead relationships across the 12-lead recording.

    \item \textbf{Clinically focused report preprocessing.}
    We refine multilingual and abbreviated ECG reports into consistent
    clinical language, suppressing routine or non-informative content and reducing ambiguity in the generation targets.

    \item \textbf{Structured clinical context-guided generation.}
    We predict diagnostic labels using a pretrained state-of-the-art classifier, utilize them to construct a clinical prompt, and
    integrate it with ECG features to provide the decoder with an explicit diagnostic plan for coherent report generation.

    \item \textbf{ECG-specific clinical evaluation metric.}
    We introduce F1-ECGBERT, a BERT-based metric that evaluates diagnostic agreement between clinically equivalent ECG reports beyond lexical similarity.

    \item \textbf{Comprehensive evaluation.}
    Extensive experiments on PTB-XL and cross-domain evaluation on MIMIC-IV-ECG show consistent improvements over state-of-the-art methods in both textual quality and diagnostic consistency.

\end{itemize}

\begin{figure*}[t]
\centering
\includegraphics[width=0.95\textwidth]{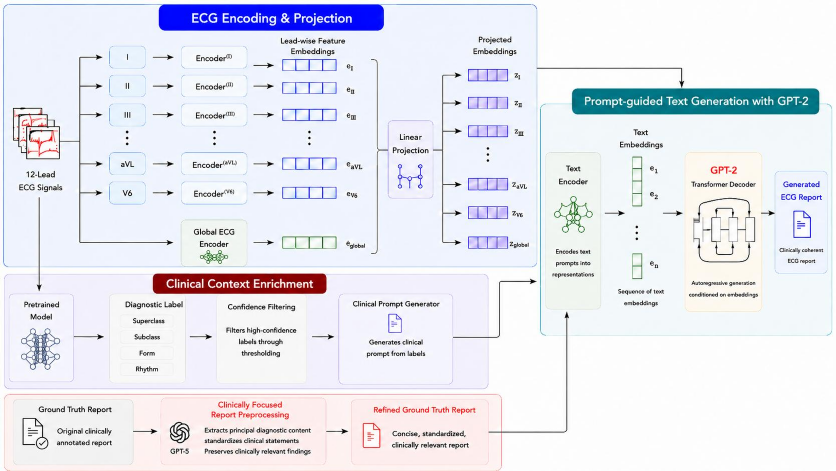}
\caption{\textbf{Overview of ECG-LENS.} Lead-specific and global encoders capture intra and inter-lead context. Predicted diagnostic labels form a structured clinical prompt that guides decoding, while LLM-refined reference reports provide clinically focused training targets.}
\label{fig:overview}
\end{figure*}

\section{Related Work}
\label{sec:related_work}
ECG classification and representation learning have advanced substantially, whereas automated ECG report generation remains comparatively underexplored. This section reviews representation-learning methods, recent ECG report-generation frameworks, and clinical evaluation metrics for assessing generated reports.



\subsection{ECG Classification and Representation Learning}
Early deep-learning approaches formulated automated ECG interpretation
primarily as a supervised classification problem, predicting one or more
diagnostic labels directly from ECG signals
\citep{hannun2019,ribeiro2020,strodthoff2021,mrml2024}. Although such models
have achieved strong performance in screening and triage, their outputs provide
limited explanation of the underlying rhythm, morphology, and lead-specific
abnormalities. Subsequent self-supervised methods learn ECG representations from unlabeled recordings, reducing the need for costly expert annotations while providing initialization for downstream tasks. ST-MEM~\citep{stmem2024} explicitly models the spatio-temporal structure of
12-lead ECG recordings through masked representation learning, whereas
HeartLang~\citep{heartlang2025} interprets heartbeats as words and rhythms as
sentences to learn representations at both morphological and rhythm levels.


Multimodal representation-learning methods leverage clinical reports to
incorporate high-level diagnostic semantics into ECG representations.
MERL~\citep{merl2024} jointly performs cross-modal ECG--report alignment and
ECG-specific uni-modal alignment, enabling zero-shot classification by matching
ECG representations with clinically enriched text prompts. D-BETA~\citep{dbeta2025} further combines masked reconstruction and contrastive
ECG--text alignment to learn transferable representations for classification. Collectively, these studies demonstrate the value of clinically informed ECG representations for downstream classification tasks. However, classification outputs are typically limited to diagnostic labels and do not provide coherent clinical narratives describing the underlying ECG findings.


\subsection{ECG Report Generation}
Recent work has begun to connect ECG representations with large language
models for automated report generation. Existing methods broadly follow two
strategies. Embedding-based approaches encode continuous ECG signals into
latent representations or pseudo-tokens that condition a language model.
MEIT~\citep{meit2025} formulates ECG reporting as multimodal instruction
following and introduces an attention-based fusion mechanism for integrating
ECG representations with multiple LLM backbones. BiECG-LLM
\citep{biecgllm2025} jointly performs classification and report generation by
fusing information from ECG signals and rendered ECG images, while
ECG-Chat~\citep{ecgchat2025} extends ECG--language interaction toward
diagnostic dialogue. Tokenization-based approaches instead convert continuous ECG recordings into
discrete symbols that can be processed more naturally by language models \cite{heartllm2026,ecgabcde2025}. 

Although these methods establish the feasibility of
generating clinical text from ECG recordings, their representations do not preserve explicit lead-specific diagnostic context or provide the decoder with a structured diagnostic plan. Effective report generation requires extracting appropriate diagnostic information from the time series data, consistent training targets, and coordinated modelling of local lead-specific and
global cardiac evidence, which is not present in the existing approaches. There has been a more significant advancement in related domains such as radiology report generation but it may not always provide useful principles for addressing these challenges because ECGs are multivariate
time-series signals whose interpretation depends on temporal morphology and
relationships across leads rather than spatial image features alone.

\subsection{Clinical Evaluation}
ECG report-generation studies commonly use BLEU, ROUGE, METEOR, and
BERTScore~\citep{zhang2020bertscore} as evaluation metrics; however, these
metrics primarily assess textual similarity. They provide only a partial
assessment of report quality and do not explicitly evaluate whether
reports preserve clinically correct diagnostic information.


In summary, existing studies have advanced ECG representation learning, report generation, and evaluation; however, these efforts remain insufficient to address the challenges of generating clinically reliable ECG reports. A comprehensive framework that integrates clinically informative ECG representations, diagnostically grounded report generation, and ECG-specific evaluation remains an open challenge.

\section{Methodology}
ECG-LENS comprises four principal components: (i) a lead-aware encoding module that combines lead-specific encoders with a global encoder that jointly models all ECG leads; (ii) a clinical context enrichment module that constructs a clinical text prompt using diagnostic labels predicted by a pretrained classifier, which supplement the ECG representations provided to the decoder; (iii) a report preprocessing scheme that produces clinically focused training targets; and (iv) a GPT-2 decoder that generates the final report. An overview of the framework is presented in Figure~\ref{fig:overview}. Rather than relying on a single innovation, ECG-LENS derives its performance from the coordinated integration of the principles introduced across these components, collectively improving the diagnostic consistency of the generated reports.

\subsection{Architecture of the Encoding Module}
\subsubsection{Input Representation}
Let an ECG recording be denoted by $X \in \mathbb{R}^{L \times T}$, where $L$ is
the number of leads and $T$ is the number of samples per lead. For a standard
12-lead ECG, $L = 12$. In the 500~Hz PTB-XL setting, each 10-second recording
contains $T = 5000$ samples per lead, giving an input of shape $12 \times 5000$.
The signal is written as
\begin{equation}
X = \{X_1, X_2, \ldots, X_L\},
\end{equation}
where $X_l$ is the time-series signal of the $l$-th lead. Because each lead
captures a different projection of cardiac electrical activity, the input
representation explicitly keeps the lead dimension rather than flattening all
leads into one sequence. The signal is normalized so that amplitude differences
caused by acquisition conditions do not dominate the representation.


\subsubsection{Lead-Specific Feature Extraction}
\label{sec:lead_specific_encoder}

ECG-LENS processes each lead independently using a one-dimensional (1D) version of the ResNet-18 encoder, rather than aggregating all leads at the input stage. The resulting feature map is temporally pooled and passed through a learnable linear projection layer:

\begin{equation}
\begin{aligned}
z_l &=
\operatorname{GAP}\!\left(
f_{\mathrm{ResNet}}^{(l)}(X_l;\theta_l)
\right)
\in \mathbb{R}^{d_r},\\
h_l &= W_p z_l + b_p
\in \mathbb{R}^{d},
\qquad l=1,2,\ldots,L.
\end{aligned}
\label{eq:lead_specific_encoding}
\end{equation}

where $f_{\mathrm{ResNet}}^{(l)}(\cdot;\theta_l)$ denotes the ResNet-18 encoder
for the $l$-th lead, $\operatorname{GAP}(\cdot)$ denotes global average pooling,
and $z_l$ is the resulting pooled representation. A Linear projection layer with parameters
$W_p \in \mathbb{R}^{d \times d_r}$ and $b_p \in \mathbb{R}^{d}$  map each
ResNet representation to the common embedding dimension $d$ required by the
subsequent fusion and generation modules.



This design prevents diagnostically relevant patterns from being obscured by
premature inter-lead aggregation. Many diagnostically important ECG patterns are localized across specific lead
groups; for example, ischemic changes associated with inferior myocardial
injury are primarily reflected in leads II, III, and aVF, whereas anterior and
lateral changes are observed predominantly in leads V1--V4 and leads I, aVL,
V5, and V6, respectively.
Preserving an explicit representation for each lead therefore allows the model to retain both localized waveform morphology and its associated lead identity. We also evaluated transformer-based encoder architectures but ResNet encoders consistently achieved stronger performance in our experiments (see Ablation Study section).

\subsubsection{Global Cardiac Context Learning}
\label{sec:global_encoder}

Although lead-specific encoders preserve localized waveform morphology, they
cannot independently model diagnostic relationships across leads. ECG-LENS
therefore employs a global one-dimensional ResNet-18 encoder that jointly
processes the complete ECG recording. For an $L$-lead ECG, the individual
signals are arranged as temporally aligned input channels:

\begin{equation*}
X_{\mathrm{ch}}
=
\operatorname{Stack}(X_1,X_2,\ldots,X_L)
\in \mathbb{R}^{L \times T},
\end{equation*}

The global encoder operates
directly on $X_{\mathrm{ch}}$, allowing its convolutional filters to jointly
capture temporal patterns and inter-lead dependencies.




\begin{equation}
\begin{aligned}
z_{\mathrm{global}}
&=
\operatorname{GAP}\!\left(
f_{\mathrm{ResNet}}^{\mathrm{global}}
(X_{\mathrm{ch}};\theta_{\mathrm{global}})
\right)
\in \mathbb{R}^{d_g},\\
h_{\mathrm{global}}
&=
W_g z_{\mathrm{global}}+b_g
\in \mathbb{R}^{d}.
\end{aligned}
\label{eq:global_encoding}
\end{equation}


The lead-specific and global representations are subsequently combined into a
single encoder output:

\begin{equation*}
H_{\mathrm{ECG}}
=
\operatorname{Stack}
\left(
h_1,h_2,\ldots,h_L,h_{\mathrm{global}}
\right)
\in \mathbb{R}^{(L+1)\times d}.
\label{eq:local_global_stack}
\end{equation*}

\begin{figure}[t]
\centering
\includegraphics[width=0.50\textwidth]{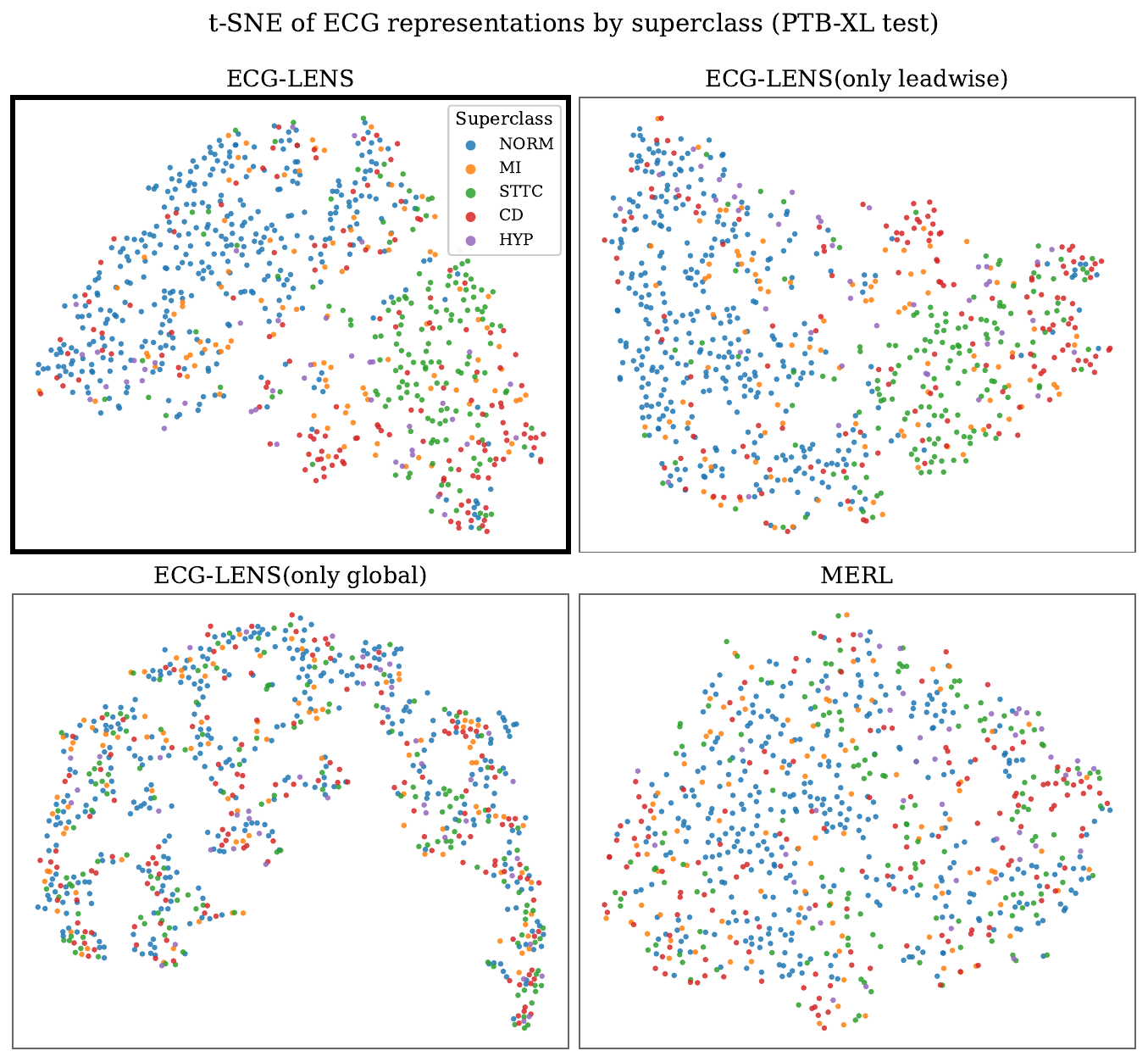}
\caption{\textbf{t-SNE plots of ECG representations by diagnostic superclasses.} ECG-LENS produces the most distinguishable clusters. }
\label{fig:tsne_leadspec}
\end{figure}

\begin{figure}[t]
\centering
\includegraphics[width=0.50\textwidth, height=100pt]{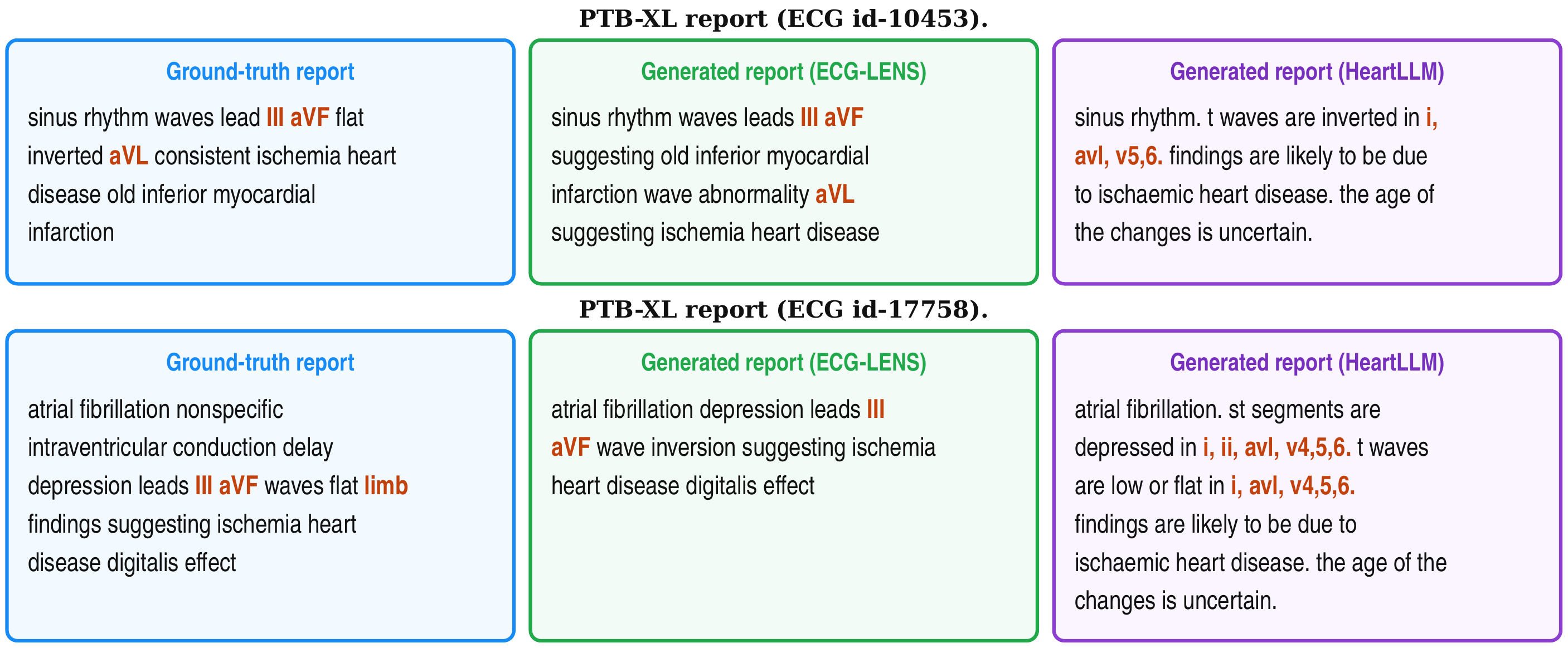}
\caption{Examples of reports demonstrating accurate lead specific abnormality detection of ECG-LENS.}
\label{fig:lead_example}
\end{figure}

Importantly, the global representation is learned directly
from the original multi-channel ECG rather than through late fusion of the
lead-specific embeddings, enabling the two encoding pathways to provide
complementary local and global cardiac information. As we can observe in Figure \ref{fig:tsne_leadspec}, the combination of global and local encoding in ECG-LENS can successfully differentiate between the various diagnostic categories, which eventually produces more clinically accurate reports. Figure~\ref{fig:lead_example} further illustrates the ability of ECG-LENS to accurately preserve lead-specific findings that competing methods may omit or misidentify.

\subsection{Clinical Context-Enriched Prompt Generation}
\label{sec:clinical_context}

The text decoder in our pipeline generates reports by jointly conditioning on ECG embeddings and a textual prompt. Rather than using a generic prompt, we provide clinical context containing preliminary diagnostic information. We first employ the publicly available highly accurate pretrained ECG classifier, MERL~\citep{merl2024}, to obtain multi-label classifiation predictions across several diagnostic categories. The resulting label confidence scores are then used to construct the clinical prompt. MERL serves as an external, frozen classifier, its parameters are not updated, and no classification or
contrastive objective is included in our training process.  
\subsubsection{Zero-Shot Classification}

MERL~\citep{merl2024} was pretrained on paired ECG recordings and clinical
reports using Cross-Modal Alignment, which aligns ECG and report
representations, together with Uni-Modal Alignment, which improves the
discriminative structure of the ECG embedding space. It performs
zero-shot classification by comparing the ECG representation with textual
embeddings of candidate diagnostic conditions and using their similarity scores
as classification confidences. Given a complete 12-lead ECG, MERL produces confidence scores
$p_c^{(k)}\in[0,1]$ for each candidate condition $c$ for each candidate condition $c$ across four diagnostic
categories $k$: superclass (common high level clinical conditions), subclass (specific diagnostic categories), rhythm, and form..
These predictions provide additional ECG-grounded guidance for the subsequent report-generation process. Notably, MERL can be replaced
with any other pretrained ECG classifier, or a diagnostic prediction head can
be attached to the ECG-LENS encoder and trained independently and used as a classifier in the pipeline, which we will explore in the future. 

\subsubsection{Prompt Construction}

ECG reports are typically concise and diagnosis-centric, with particular
conditions frequently expressed through recurring clinical phrases
\citep{meit2025}. Rather than conditioning the GPT-2 decoder on a generic
instruction , we transform reliable MERL
predictions into a compact paragraph that provides preliminary
diagnostic context.

Only high-confidence predictions are included in the prompt. Because the appropriate confidence cutoff may differ across superclass, subclass, rhythm,
and form categories, we determine category-specific thresholds using metric specific optimization using grid search in the
validation set, which are applied
during testing and cross-domain evaluation.


Each retained condition is mapped to a normalized, condition-specific phrase
commonly used in ECG interpretation following the international SCP-ECG standard by the prompt-construction function and then organized into a short and
coherent clinical paragraph. For example, high-confidence predictions of myocardial infarction
(superclass), inferior myocardial infarction (subclass), non-specific ST
elevation (form), and atrial fibrillation (rhythm) may produce the following
prompt:

\begin{quote}
\small
\texttt{The ECG is suggestive of inferior myocardial infarction.
Non-specific ST-segment elevation is present, and atrial fibrillation is also
indicated.}
\end{quote}

Conditioning the text decoder on a prompt with diagnostic context significantly increases the report quality. Consequently, the decoder receives both direct morphological evidence
from the ECG encoders and explicit contextual guidance regarding the most
probable conditions, improving the diagnostic grounding and overall quality of
the generated report. Examples of some prompts and results regarding performance gain are presented in Supplementary Section 4.

\subsection{Report Generation Using Text Decoder}
\label{sec:report_generation}

ECG-LENS uses a lightweight GPT-2--style autoregressive decoder
\citep{radford2019gpt2} trained from scratch, without initialization from
pretrained GPT-2 weights. To represent the specialized and abbreviation-rich
language of ECG interpretations, we construct a domain-specific tokenizer from
the complete collection of preprocessed PTB-XL reports. The resulting ECG
vocabulary is used to tokenize both the clinical prompts and target reports,
allowing the decoder to represent recurring diagnostic terms and
condition-specific phrases consistently.

Let $\mathcal{T}_{\mathrm{ECG}}$ denote the custom tokenizer and
$R=(w_1,w_2,\ldots,w_N)=\mathcal{T}_{\mathrm{ECG}}(R_{\mathrm{text}})$ denote a
tokenized target report. Given the combined local--global ECG representation
$H_{\mathrm{ECG}}$ and clinical prompt $p$, the decoder models

\begin{equation}
P_{\Theta}(R\mid X,p)
=
\prod_{t=1}^{N}
P_{\Theta}
\left(
w_t
\mid
w_{<t},H_{\mathrm{ECG}},p
\right)
\end{equation}

Thus, each generated token is conditioned on the preceding report tokens,
lead-specific and global ECG features, and the preliminary diagnostic context
provided by the prompt.

ECG-LENS is optimized in a single end-to-end training setup. Let
$
\Theta=
\left\{
\theta_{\mathrm{lead}},
\theta_{\mathrm{global}},
\theta_{\mathrm{proj}},
\theta_{\mathrm{dec}}
\right\}$ denote the trainable parameters of the lead-specific encoders, global encoder, linear projection layers, and text decoder, respectively. For each training
sample $(X^{(i)},R^{(i)})$, we generate the clinical prompt using the frozen MERL model with parameters
$\phi_{\mathrm{MERL}}$ during the forward pass: $p^{(i)}
=
g_{\mathrm{prompt}}
\left(
f_{\mathrm{MERL}}
\left(
X^{(i)};\phi_{\mathrm{MERL}}
\right)
\right)$. The complete trainable model minimizes the autoregressive negative
log-likelihood of the target reports:

\begin{equation}
\mathcal{L}_{\mathrm{gen}}(\Theta)
=
-\frac{1}{M}
\sum_{i=1}^{M}
\sum_{t=1}^{N_i}
\log
P_{\Theta}
\left(
w_t^{(i)}
\mid
w_{<t}^{(i)},
H_{\mathrm{ECG}}^{(i)},
p^{(i)}
\right)
\end{equation}

Gradients are propagated jointly through the decoder, projection layers, and
both ECG encoding pathways. In contrast, MERL participates only in the forward
pass for prompt generation and remains excluded from backpropagation.



\subsection{Clinically Focused Report Preprocessing}
\label{sec:report_preprocessing}

Raw ECG reports frequently contain repetitive, non-informative, or record-specific content that does not generalize across patients. Examples
include isolated voltage measurements, irrelevant patient information,
administrative text, redundant statements, and routine phrases that provide
little diagnostic value. Training directly on such reports can introduce
spurious variation into the generation targets and distract the model from
clinically meaningful findings.

We therefore employ a few-shot prompting strategy with GPT-5.5 to extract the
principal diagnostic content from each report. The prompt explicitly lists and instructs on statements to discard and clinically important information to preserve. and  contains several
representative pairs of original and refined reports, demonstrating how to
remove irrelevant information while preserving rhythm, morphology, diagnostic
conditions, finding--lead associations, and other clinically significant
observations. The refinement process is explicitly instructed not to introduce,
remove, or alter any diagnosis supported by the original report. Moreover, PTB-XL is a multilingual dataset and we use this process to translate the reports using standardized clinical statements.

The resulting reports are reviewed by a trained cardiologist to verify that the clinically
relevant findings are preserved and that no unsupported information is
introduced. These concise and standardized reports are subsequently used as
the generation targets for ECG-LENS. Further details and the prompt used are shown in Supplementary Section 3. 



\subsection{ECG-Specific Clinical Evaluation}
\label{sec:f1_ecgbert}

\begin{figure}[t]
\centering
\includegraphics[width=\columnwidth]{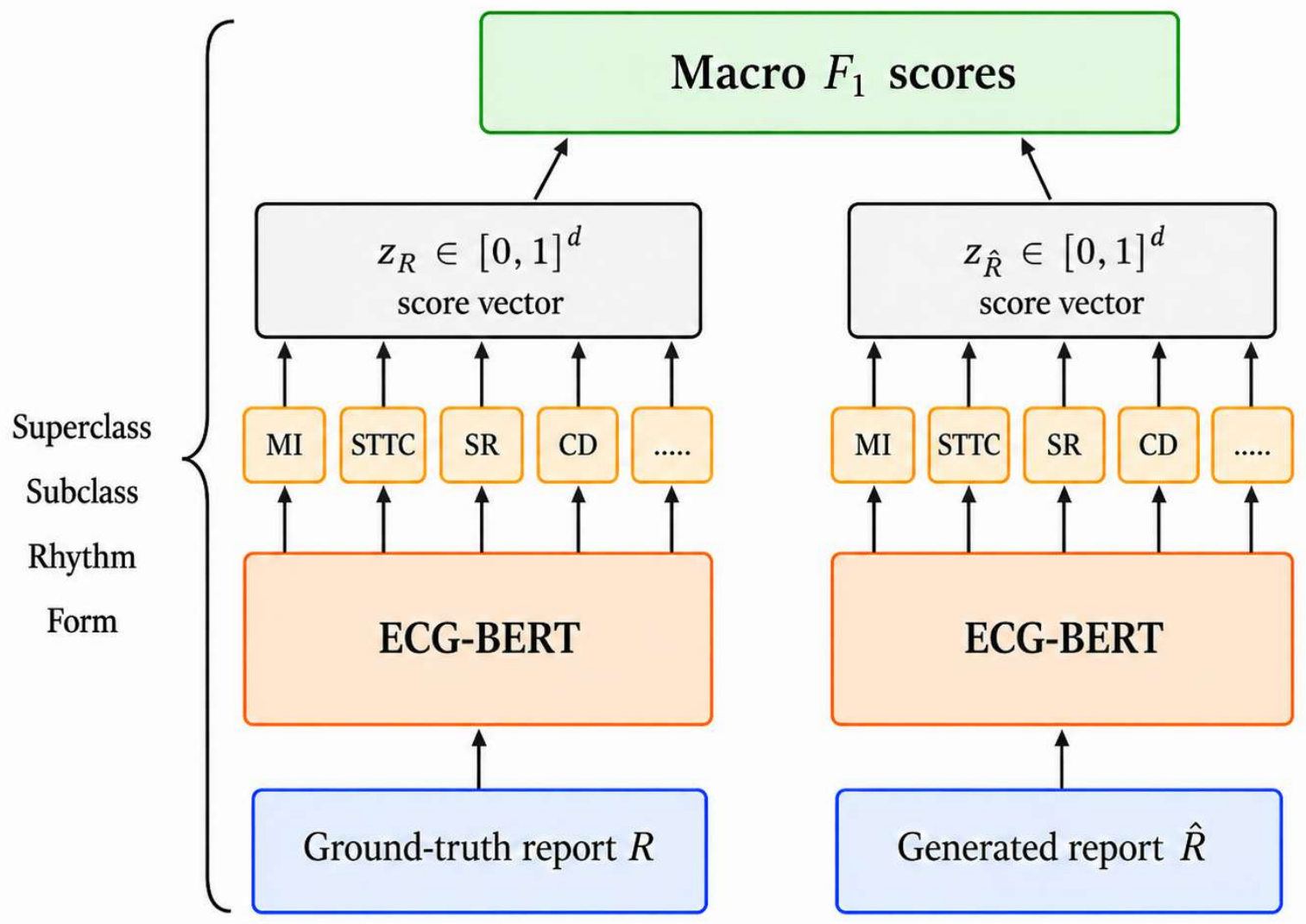}
\caption{\textbf{F1-ECGBERT evaluation.} The reference report $R$ and the generated
report $\hat{R}$ are passed through the BERT evaluator (one per
diagnostic category) to extract the appropriate labels for the reports.$F_1$ score between the classification results measures whether the generated report preserves
the diagnostic content of the reference.}
\label{fig:evalmodel}
\end{figure}

Conventional lexical metrics such as BLEU, ROUGE, and METEOR primarily measure
word or phrase overlap and may therefore fail to assess the clinical
consistency of a generated ECG report. We therefore introduce F1-ECGBERT, a BERT-based evaluation metric that measures agreement between the diagnostic
content of generated and reference reports, inspired by ChexBert \cite{chexbert2020} metric used in radiology. As illustrated in Figure~\ref{fig:evalmodel}, we train four independent
report-to-label BERT classifiers corresponding to the diagnostic
superclass, diagnostic subclass, rhythm, and form categories. Each model takes a text report as input and predicts the probabilities of the conditions
belonging to its diagnostic category. The four classifiers are trained separately using binary cross-entropy and remain fixed during report evaluation.

For each diagnostic category, the corresponding classifier extracts binary
label vectors from both the reference report $R$ and generated report
$\hat{R}$. Their agreement is measured by calculating the F1-score. A high F1-ECGBERT score indicates that the generated report preserves the
diagnostic conditions expressed in the reference report, providing a more
clinically meaningful assessment than lexical similarity alone. We perform rigorous validation of the proposed metric on an extensive set of example reports, demonstrating its effectiveness in capturing ECG semantics and superiority over conventional metrics. Details are provided in Supplementary Section 1.

\section{Experiments \& Results}

\subsection{Experiment Setup}
\subsubsection{Datasets}
We use \textbf{PTB-XL}~\citep{ptbxl2020} for training and experiments, which provides 21,799 10-second 12-lead recordings from 18,869  patients. We
use the 500~Hz signals ($12\times5000$) and the official \texttt{strat\_fold}
split (folds 1--8 train, 9 validation, 10 test). Then, \textbf{MIMIC-IV-ECG}%
~\citep{mimicivecg2023} is used for cross-domain evaluation, that is to test whether the model learns robust ECG--report relationships
rather than PTB-XL-specific wording.

\subsubsection{Baselines}
We compare against ECG report generation methods MEIT~\citep{meit2025},
ECG-aBcDe~\citep{ecgabcde2025}, BiECG-LLM~\citep{biecgllm2025},
ECG-Chat~\citep{ecgchat2025}, and HeartLLM~\citep{heartllm2026}.

\subsubsection{Evaluation Metrics}
We use widely used NLP metrics BLEU~\citep{papineni2002bleu}, ROUGE~\citep{lin2004rouge}, METEOR~\citep{banerjee2005meteor}, and our newly defined F1-ECGBERT over the superclass,
subclass, rhythm, and form categories. 

\subsubsection{Implementation Details}
Each of the lead-specific encoders and the
global encoder is a 1-D ResNet-18 (convolution kernels $[9,7,7,5]$, stem kernel
$9$) with embedding dimension $D{=}512$. The
decoder is a GPT-2--style transformer trained from scratch with $12$ layers, $8$
attention heads, hidden size $512$, and dropout $0.5$, over a domain vocabulary
built from the preprocessed reports. MERL remains frozen and used for classification. 

We minimize the report-generation loss with Adam (encoder
learning rate $4{\times}10^{-4}$, decoder $1{\times}10^{-4}$), batch size $32$,
gradient clipping $5$, and learning-rate decay $\times0.8$ after $8$ epochs
without improvement, for up to $200$ epochs with early stopping (patience $30$).
Models are implemented in PyTorch and trained on a single NVIDIA GeForce RTX 4090 GPU with 24 GB of memory.

\subsection{Results}


\subsubsection{In-Domain Results on PTB-XL}
\label{sec:ptbxl_results}

We first test ECG-LENS and the baselines in PTB-XL official test split. Table~\ref{tab:ptbxl} compares
ECG-LENS with existing ECG report-generation methods using standard NLP metrics.

\begin{table}[t]
\centering
\small
\setlength{\tabcolsep}{2.1pt}
\begin{tabular}{lcccccc}
\toprule
\textbf{Method} &
\textbf{B-1} &
\textbf{B-2} &
\textbf{B-3} &
\textbf{R-1} &
\textbf{R-L} &
\textbf{MET} \\
\midrule
MEIT         & 0.371 & 0.356 & 0.343 & 0.557 & 0.518 & 0.490 \\
ECG-aBcDe    & 0.388 & 0.365 & --    & --    & 0.521 & 0.421 \\
BiECG-LLM    & \cellcolor{gray!25}0.521
             & \cellcolor{gray!25}0.441
             & --
             & \cellcolor{gray!25}0.713
             & \cellcolor{gray!25}0.601
             & \cellcolor{gray!25}0.682 \\
HeartLLM     & 0.496 & 0.434 & \cellcolor{gray!25}0.375 & 0.595 & 0.592 & 0.539 \\
\textbf{ECG-LENS}
             & \textbf{0.651}
             & \textbf{0.569}
             & \textbf{0.501}
             & \textbf{0.729}
             & \textbf{0.686}
             & \textbf{0.714} \\
\bottomrule
\end{tabular}
\caption{PTB-XL in-domain text-generation results. B, R, and MET denote
BLEU, ROUGE, and METEOR, respectively. Higher values are better. A dash denotes a
result that was not reported in the original publication and could not be
reproduced because no public implementation was available. Bold indicates the best result, gray shading indicates the second-best result.}
\label{tab:ptbxl}
\end{table}

ECG-LENS achieves the highest score across every reported metric.
Significant improvements over BLEU scores, specifically absolute
gains of 12.8\% in BLEU-2 and 12.6\% in BLEU-3 over the second best method, demonstrates good lexical overlap in the generated reports. It also improves ROUGE-L and METEOR by 8.5\% and 3.2\%,
respectively, indicating stronger structural alignment and semantic consistency
with the reference reports.

We further evaluate diagnostic consistency using the four independently
trained F1-ECGBERT evaluators (only for the methods with publicly available implementations). As shown in
Table~\ref{tab:ecgbert}, ECG-LENS obtains the highest score for every
diagnostic category. Specifically, ECG-LENS exceeds the strongest baseline by 10.9\% for diagnostic superclass category. Consistent
improvements in diagnostic subclass and rhythm further demonstrate its ability to preserve clinically relevant findings.

\begin{table}[t]
\centering
\small
\setlength{\tabcolsep}{2.5pt}
\begin{tabular}{lcccc}
\toprule
\textbf{Method} &
\textbf{Superclass} &
\textbf{Subclass} &
\textbf{Rhythm} &
\textbf{Form} \\
\midrule
ECG-Chat     & 0.653 & \cellcolor{gray!25}0.678 & 0.635 & 0.613 \\
HeartLLM     & \cellcolor{gray!25}0.659 & 0.669 & \cellcolor{gray!25}0.667 & \cellcolor{gray!25}0.689 \\
MEIT         & 0.523 & 0.587 & 0.556 & 0.532 \\
\textbf{ECG-LENS}
             & \textbf{0.768}
             & \textbf{0.745}
             & \textbf{0.732}
             & \textbf{0.702} \\
\bottomrule
\end{tabular}
\caption{F1-ECGBERT scores on the PTB-XL test set across four diagnostic
categories.}
\label{tab:ecgbert}
\end{table}

\begin{table}[t]
\centering
\small
\setlength{\tabcolsep}{2.1pt}
\begin{tabular}{lcccccc}
\toprule
\textbf{Method} &
\textbf{B-1} &
\textbf{B-2} &
\textbf{B-3} &
\textbf{R-1} &
\textbf{R-L} &
\textbf{MET} \\
\midrule
MEIT         & 0.352 & 0.331 & 0.281 & 0.521 & 0.496 & 0.472 \\
ECG-aBcDe    & 0.388 & 0.365 & 0.314 & 0.548 & 0.521 & 0.421 \\
BiECG-LLM    & \cellcolor{gray!25}0.506
             & \cellcolor{gray!25}0.429
             & \cellcolor{gray!25}0.367
             & \cellcolor{gray!25}0.618
             & \cellcolor{gray!25}0.589
             & \cellcolor{gray!25}0.661 \\
\textbf{ECG-LENS}
             & \textbf{0.628}
             & \textbf{0.562}
             & \textbf{0.472}
             & \textbf{0.634}
             & \textbf{0.612}
             & \textbf{0.701} \\
\bottomrule
\end{tabular}
\caption{Cross-domain evaluation on MIMIC-IV-ECG.}
\label{tab:mimic}
\end{table}

ECG-LENS maintains the highest performance across all six metrics under
cross-domain evaluation and achieves comparable scores in all metrics, with respect to PTB-XL instead of domain shift. Performance gain of 4\% in METEOR and even larger gain of more than 10\% in BLEU scores proves that ECG-LENS generalizes the ECG signal characteristics, and it both learns ECG report phrases and
stronger semantic correspondence with the reference reports.


\subsubsection{Cardiologist Evaluation and Qualitative Analysis}
\label{sec:qualitative_analysis}

A randomly selected subset of the generated reports was reviewed by a trained cardiologist, who judged 67\% to be fully clinically correct and nearly all remaining reports to be partially correct. Further details are provided in Supplementary Section 5. We further qualitatively examine the generated reports against ground truth reports for preserved diagnoses, lead-specific findings, rare disease identification, and missing abnormalities. ECG-LENS successfully captures the conditions in most of the cases and successfully identify rare cases in a lot of scenarios. Representative comparisons with reference reports,
together with a detailed analysis of common failure modes, are provided in Supplementary Section 2. 

\subsubsection{Ablation Study}
\label{sec:ablation}

We evaluate five configurations under identical data splits, tokenizer,
decoder capacity, and optimization settings. Rows (a) and (b) compare the ResNet and Transformer global encoders, whereas rows (c)-(e) trace the progressive development of ECG-LENS by successively incorporating report preprocessing, lead-specific encoding, and clinical prompting into the ResNet baseline.

\begin{table}[t]
\centering
\footnotesize
\setlength{\tabcolsep}{2.5pt}
\renewcommand{\arraystretch}{1.08}
\begin{tabular}{@{}p{2.55cm}cccccc@{}}
\toprule
\textbf{Configuration} &
\textbf{B-1} & \textbf{B-2} & \textbf{B-3} &
\textbf{R-1} & \textbf{R-L} & \textbf{MET} \\
\midrule
(a) Global encoding only (ResNet-1D)
& 0.475 & 0.392 & 0.328 & 0.588 & 0.572 & 0.556 \\

(b) Global encoding only (Transformer)
& 0.462 & 0.388 & 0.318 & 0.578 & 0.552 & 0.543 \\

(c) (a) + report preprocessing
& 0.558 & 0.468 & 0.398 & 0.617 & 0.625 & 0.598 \\

(d) (c) + lead-specific encoders
& 0.572 & 0.502 & 0.437 & 0.684 & 0.652 & 0.650 \\

\textbf{(e) Final model ((d) + clinical prompt)}
& \textbf{0.651} & \textbf{0.569} & \textbf{0.501} &
\textbf{0.729} & \textbf{0.686} & \textbf{0.714} \\
\bottomrule
\end{tabular}
\caption{Ablation study on the PTB-XL test set. Rows (a)-(b) compare
encoder alternatives, while rows (c)-(e) progressively incorporate the
proposed components.}
\label{tab:ablation}
\end{table}

The ResNet global encoder consistently outperforms its Transformer counterpart and is therefore adopted in all subsequent stages of the study. Each subsequent component provides significant further gains across almost all metrics.

\section{Conclusion}

We introduced ECG-LENS, an end-to-end ECG report-generation framework that integrates lead-specific and global signal encoding, clinically enriched prompting, focused report preprocessing, and autoregressive text generation. These complementary innovations span the complete ECG-to-report pipeline and collectively improve both the linguistic quality and clinical reliability of generated reports. Across in-domain and cross-domain evaluations, ECG-LENS achieves improvements of 4\%, 6.3\%, and 11.5\% in METEOR, ROUGE-L, and F1-ECGBERT, respectively, over the strongest corresponding baselines. We also introduce F1-ECGBERT, an ECG-specific evaluation metric that assesses diagnostic agreement beyond surface-level lexical similarity. Moreover, our proposed model is lightweight with very fast inference time of around 30ms, making it suitable for resource constrained settings. 

Despite these promising results, some limitations remain.  First, computational constraints prevented a systematic investigation of larger-capacity encoders and decoders, which may improve representation learning and report quality. Second, although ECG-LENS substantially reduces clinical inconsistencies, it still occasionally hallucinates or makes mistakes in a small number of cases, as illustrated in Supplementary Section 2. Therefore, future work should explore larger architectures and hallucination-mitigation strategies. Overall, ECG-LENS represents a significant step toward reliable and clinically grounded automated ECG report generation.


\bibliography{aaai2027}
\end{document}